\documentclass[sigconf]{acmart}

\usepackage{mmstyles}
\usepackage{bm}

\definecolor{OURS}{RGB}{90, 180, 172}
\definecolor{UGCN}{RGB}{216, 179, 101}
\definecolor{GT}{RGB}{160, 160, 160}

\AtBeginDocument{%
  \providecommand\BibTeX{{%
    \normalfont B\kern-0.5em{\scshape i\kern-0.25em b}\kern-0.8em\TeX}}}

\copyrightyear{2021}
\acmYear{2021}
\setcopyright{acmcopyright}\acmConference[MM '21]{Proceedings of the 29th ACM International Conference on Multimedia}{October 20--24, 2021}{Virtual Event, China}
\acmBooktitle{Proceedings of the 29th ACM International Conference on Multimedia (MM '21), October 20--24, 2021, Virtual Event, China}
\acmPrice{15.00}
\acmDOI{10.1145/3474085.3475219}
\acmISBN{978-1-4503-8651-7/21/10}

\begin{document}
\fancyhead{}
	
\title{Conditional Directed Graph Convolution \\ for 3D Human Pose Estimation}

\renewcommand\footnoterule{\kern-3pt \hrule width 2in \kern 2.6pt}

\renewcommand{\thefootnote}{\fnsymbol{footnote}}
\author{Wenbo Hu$^{1, 2, *}$, Changgong Zhang$^{2, *}$, Fangneng Zhan$^{3}$, Lei Zhang$^{2, 4}$, Tien-Tsin Wong$^{1, \dagger}$}
\affiliation{%
\institution{$^{1}$ The Chinese University of Hong Kong \ \ $^{2}$ DAMO Academy, Alibaba Group}
    \country{}
}
\affiliation{\institution{$^{3}$ Nanyang Technological University \ \ $^{4}$ The Hong Kong Polytechnic University}
    \country{}
}
\affiliation{\{wbhu, ttwong\}@cse.cuhk.edu.hk, changgong.zcg@alibaba-inc.com, fnzhan@ntu.edu.sg, cslzhang@comp.polyu.edu.hk
    \country{}
}
\def\authors{Wenbo Hu, Changgong Zhang, Fangneng Zhan, Lei Zhang, Tien-Tsin Wong}
\renewcommand{\shortauthors}{Hu et al.}
	
	
	\begin{abstract}
			\footnotetext{$^{*}$Equal contribution \quad $^{\dagger}$Corresponding author} 
		Graph convolutional networks have significantly improved 3D human pose estimation by representing the human skeleton as an undirected graph.
		However, this representation fails to reflect the articulated characteristic of human skeletons as the hierarchical orders among the joints are not explicitly presented.
		In this paper, we propose to represent the human skeleton as a \emph{directed} graph with the joints as nodes and bones as edges that are directed from parent joints to child joints.
		By so doing, the directions of edges can explicitly reflect the hierarchical relationships among the nodes.
		Based on this representation,		
		we further propose a spatial-temporal \emph{conditional} directed graph convolution to leverage varying non-local dependence for different poses by conditioning the graph topology on input poses. 
		Altogether, we form a U-shaped network,
		named \emph{U-shaped Conditional Directed Graph Convolutional Network}, for 3D human pose estimation from monocular videos.
		To evaluate the effectiveness of our method, we conducted extensive experiments on two challenging large-scale benchmarks: Human3.6M and MPI-INF-3DHP.
		Both quantitative and qualitative results show that our method achieves top performance.	
		Also, ablation studies show that directed graphs can better exploit the hierarchy of articulated human skeletons than undirected graphs, and the conditional connections can yield adaptive graph topologies for different poses.

	\end{abstract}

	
	
	\begin{teaserfigure}
		\centering
		\vspace*{-2mm}
		\includegraphics[width=0.91\textwidth]{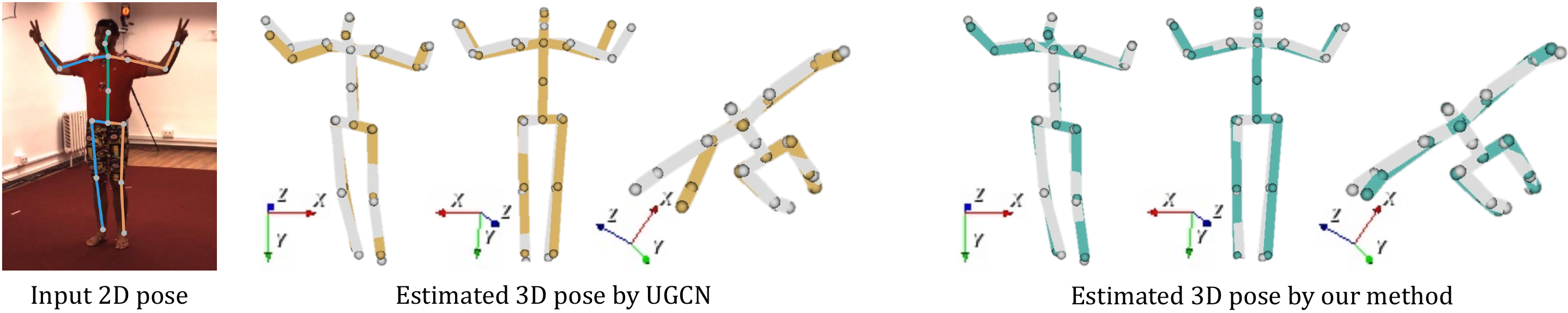}
		\vspace*{-4mm}
		\caption{Given a sequence of 2D human poses estimated by an off-the-shelf 2D pose estimator, \eg, HR-Net~\cite{sun2019deep}, \textcolor{OURS}{our method} can produce more precise 3D poses compared with state-of-the-art approach, \textcolor{UGCN}{UGCN}~\cite{wang2020motion}.
		We show the results under three different viewpoints as indicated by the 3D orientation markers.
		And ground-truth 3D poses are shown in \textcolor{GT}{gray} as a reference.
		}
		\vspace*{1mm}
		\label{fig:teaser}
	\end{teaserfigure}

	\maketitle
	
\section{Introduction}
\label{sec:introduction}

Human pose estimation from monocular videos plays a critical role in a wide spectrum of applications, \eg, action recognition~\cite{luvizon20182d,yan2018spatial}, athletic training~\cite{wang2019ai}, data-driven computer animation, and gaming.
Compared with the 2D pose in image space, the 3D pose in physical space is more informative. However, estimating 3D poses from monocular videos is much more challenging due to the depth ambiguity, metric inconsistency (\ie, millimeters instead of pixels), and the high non-linearity of human dynamics. 
Given a monocular video of human motions acquired, for example, from consumer-level cameras, our ultimate goal is to estimate the human pose sequence in the 3D physical space.
Following~\cite{martinez2017simple,pavllo20193d,wang2020motion}, we define the 3D pose as the 3D locations of joints, including \textit{``head'', ``shoulders'', ``knees'', ``elbows'',} and so no, of the human body.

Thanks to the development of deep learning, we have witnessed remarkable achievements in 3D pose reasoning~\cite{martinez2017simple,pavllo20193d,cai2019exploiting,voxelpose,wang2020motion,liu2020comprehensive,iqbal2020weakly,cheng2019occlusion,xu2020deep,cheng20203d,shi2020motionet,liu2020attention}.
State-of-the-art approaches~\cite{pavllo20193d,cheng2019occlusion,cai2019exploiting,wang2020motion,cheng20203d,liu2020attention} typically divide the task into two stages: 2D pose estimation to localize the keypoints in the image space, and predicting joint positions in the 3D space from 2D pixel coordinates.
We follow this strategy and focus on the second stage, lifting 2D pixel coordinates to 3D positions,
while the first stage, 2D pose estimation, is also a popular vision task that is quite well-studied in~\cite{chen2018cascaded,sun2019deep,cao2019openpose}.
Recent efforts~\cite{cai2019exploiting,wang2020motion} have explored to represent the human pose as an undirected spatial-temporal graph and thereafter employ graph convolution networks to estimate the 3D pose.
Compared with representing human pose as a time sequence of independent joint location vectors~\cite{martinez2017simple,pavllo20193d,cheng2019occlusion,liu2020attention}, the undirected graph representation is more relevant to the inherent nature of human skeletons.

\begin{figure}[!t] 
	\centering
	\includegraphics[width=0.95\linewidth]{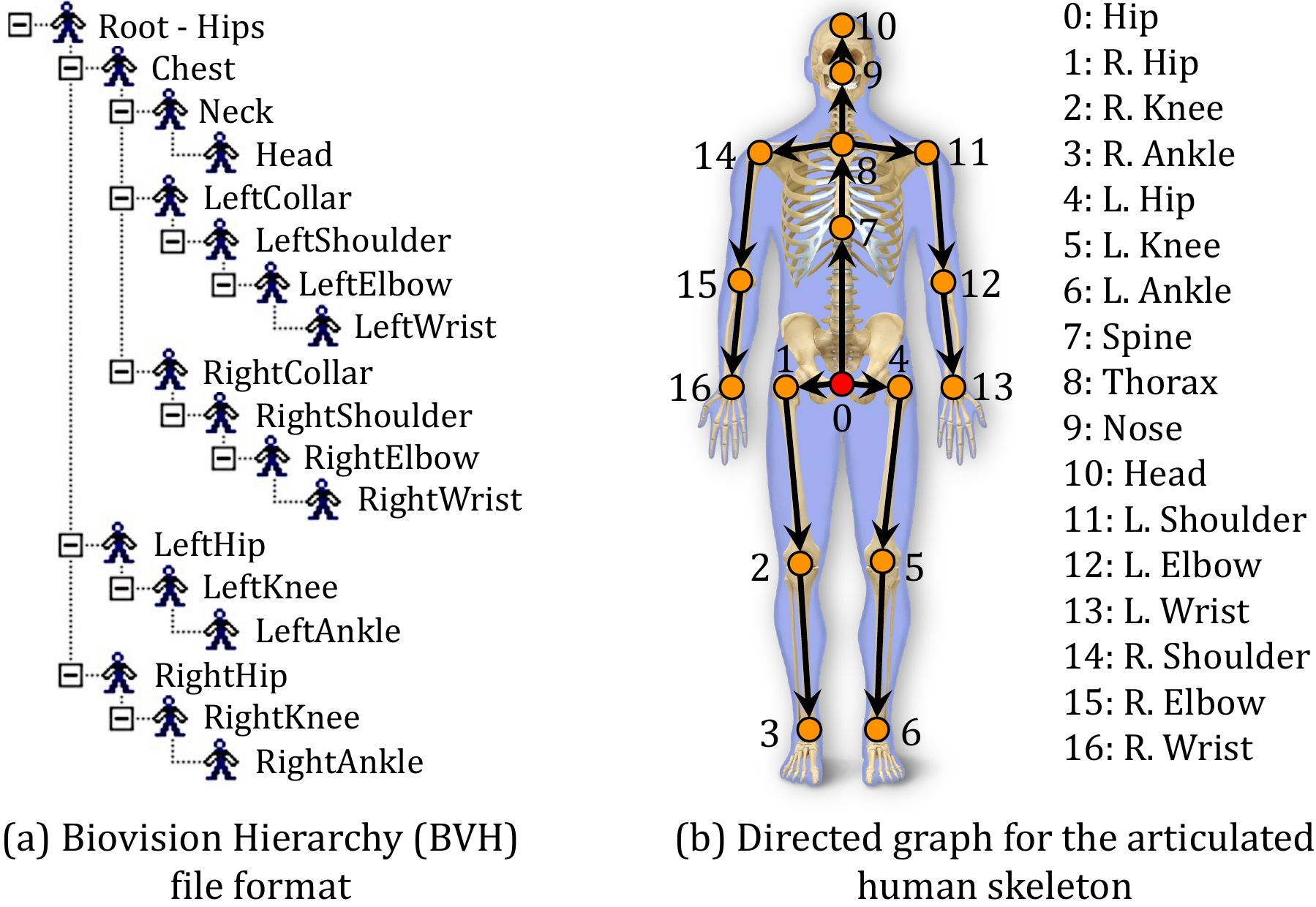}
	\vspace{-2mm}
	\caption{
		The BVH file format (a) is a concrete representation of the hierarchical structure of human skeletons. We adopt the directed graph (b) to represent the articulated human skeleton, with the hierarchy is represented by the directions of edges. The ``hip'' joint (red dot in (b)) is set as the  directed graph's root node.
	}
	\vspace{-5mm}
		\label{fig: BVH}
\end{figure}

However, the undirected graph representation does not take into account the hierarchical structure of bones, which is one of the most significant characteristics of human anatomy.
%
For example, when a person moves the shoulder joints, the following joints (\ie elbows and wrists) are moved as well according to the anatomical articulation while not the other way around.
%
This hierarchy is widely utilized to represent human motion in computer animation as evidenced by the Biovision Hierarchy (BVH) file format,
which presents a typical hierarchical structure as illustrated in Figure~\ref{fig: BVH} (a).
%
To this end,
we propose to represent the human skeleton as a \emph{directed} graph with the joints as nodes and bones as edges that are directed from parent joints to child joints, as shown in Figure~\ref{fig: BVH} (b).
By doing so, the hierarchical relationships among all the joints are explicitly presented by directions of the edges.
%
%

\begin{figure}[!t] 
	\centering
	\includegraphics[width=0.92\linewidth]{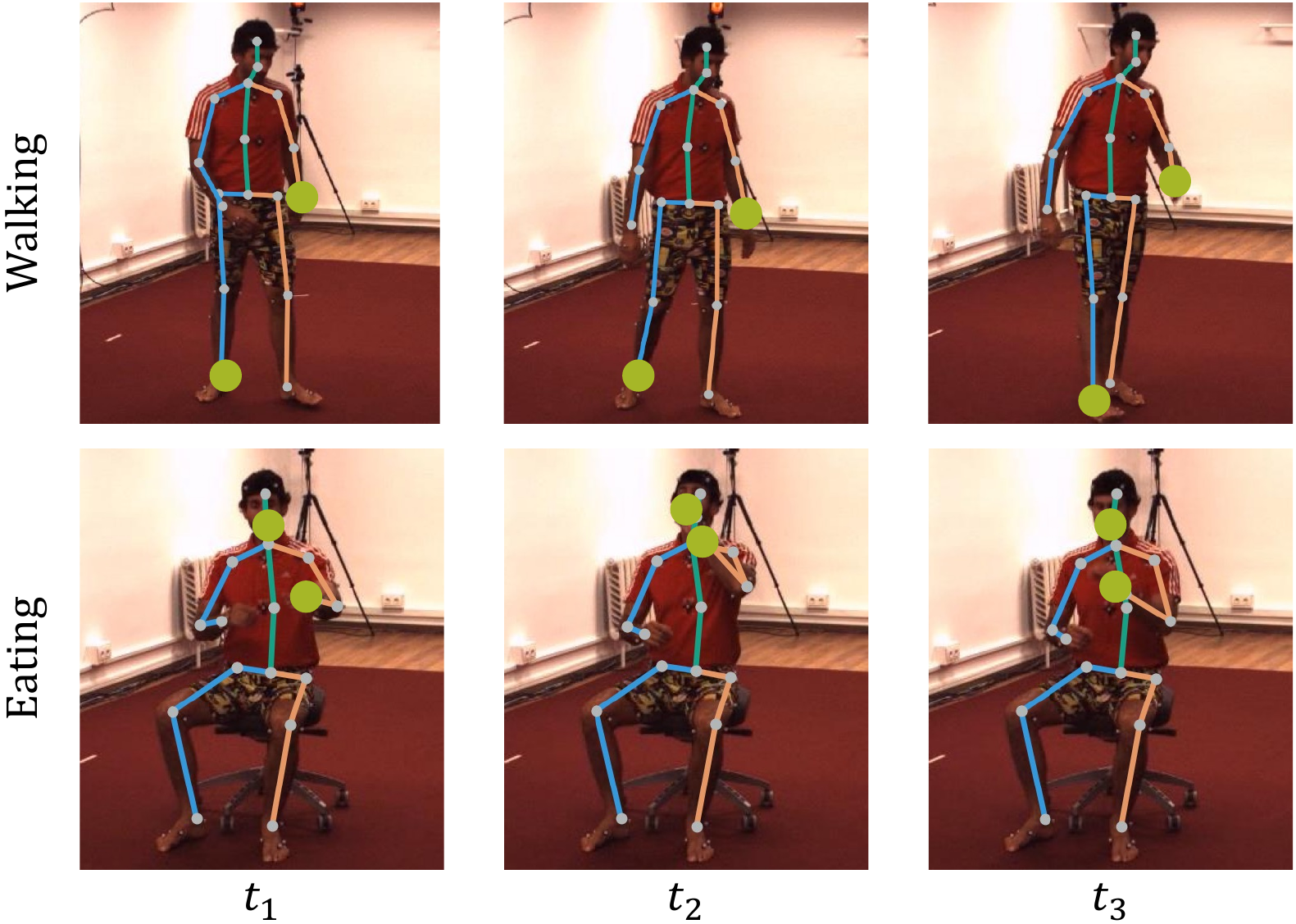}
	\vspace{-2mm}
	\caption{
		Varying non-local dependence among joints. Dependence between {``left hand''} and {``right foot''} is significant when people are walking (first row) since this pose can help to keep balance, while dependence between {``left hand''} and {``head''} is high when people are eating (second row). $t_1$, $t_2$, and $t_3$ are time indices.
	}
	\vspace{-5mm}
	\label{fig: conditional_connections}
\end{figure} 

After representing 2D poses as a sequence of directed graphs, we can employ the spatial-temporal directed graph convolution (ST-DGConv) to extract features.
However, ST-DGConv shares the same graph topology among all kinds of poses, which may not be optimal since there is non-local dependence among the nodes and the non-local dependence varies a lot for different poses.
%
%
%
For example, as shown in Figure~\ref{fig: conditional_connections}, the dependence between \textit{``left hand''} and \textit{``right foot''} joints is obviously significant when people are walking (since this pose can help keep balance). In contrast, the dependence between \textit{``hands''} and \textit{``head''} joints would be high when eating.
Inspired by the conditional convolution~\cite{NEURIPS2019_f2201f51,tian2020conditional} that enables different data samples using different convolution kernels, we propose a \emph{spatial-temporal conditional directed graph convolution} (ST-CondDGConv) to condition the connections of the directed graph on input poses, such that different kinds of poses can adopt appropriate connections to optimally leverage non-local dependence.
To the best of our knowledge, this is the first attempt to introduce conditional connections to directed graph convolution for 3D human pose estimation.
Overall, we composite a U-shaped network with the ST-DGConv and ST-CondDGConv layers, named \emph{U-shaped Conditional Directed Graph Convolutional Network} (U-CondDGCN), to capture temporal relationship in both short temporal intervals and long temporal ranges.

We evaluated our model on two large-scale 3D human pose estimation benchmarks: Human3.6M~\cite{ionescu2013human3} and MPI-INF-3DHP~\cite{mehta2017monocular}.
Both quantitative and qualitative experimental results show our method achieves top performance.
Moreover, we conducted extensive ablation studies to demonstrate that the directed graph can better utilize the hierarchy of articulated human skeletons, and the conditional connections can yield adaptive graph topologies for different kinds of poses.
Our contributions are summarized below.

\begin{itemize}
	\item We argue that the hierarchy of articulated skeletons is beneficial for 3D pose reasoning, and directed graphs are more suitable to model the hierarchy than undirected graphs. 
	
	\item We propose a novel conditional directed graph convolution to enable adaptive graph topologies for different pose samples at inference time, such that different poses can benefit from appropriate non-local dependence.
	
	\item We present a U-shaped Conditional Directed Graph Convolutional Network (U-CondDGCN) for 3D human pose estimation from detected 2D keypoints. Our method achieves top performance on the challenging Human3.6M and MPI-INF-3DHP benchmarks, thus demonstrating its effectiveness.
\end{itemize}


\begin{figure*}[!t] 
	\centering
	\includegraphics[width=0.99\linewidth]{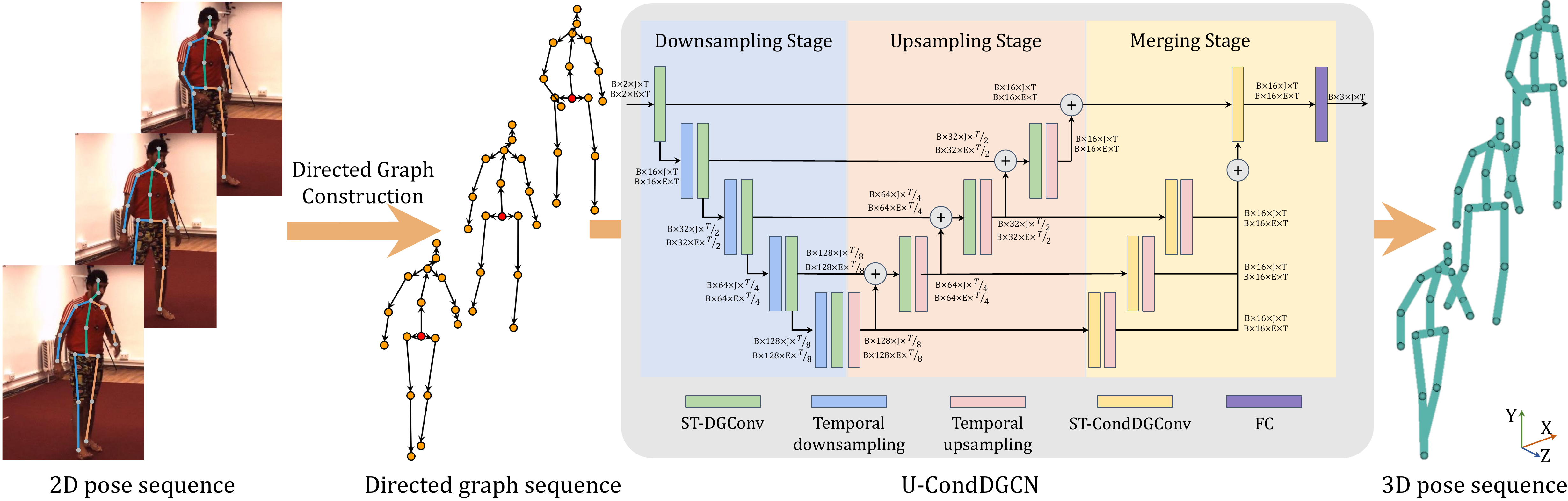}
	\vspace{-3mm}
	\caption{
		Overview of our framework. Given a sequence of 2D poses estimated by any off-the-shelf 2D pose estimators, we first construct a sequence of directed graphs and then estimate the 3D poses with our U-shaped conditional directed graph convolutional network (U-CondDGCN).
	}
	\vspace{-4mm}
	\label{fig: overview}
\end{figure*} 

\section{Related Work}
\label{sec:relatedWork}

\subsection{3D Human Pose Estimation}
\label{sec:3dhpe}
With the development of deep learning, 2D human pose estimation~\cite{newell2016stacked,chen2018cascaded,sun2019deep,cao2019openpose} has shown remarkable progress, while 3D pose estimation remains more challenging due to the depth ambiguity.
Several methods~\cite{liang2019shape,qiu2019cross,iskakov2019learnable,he2020epipolar,voxelpose} propose to relieve this issue by adopting multi-view images/videos as input.
However, multi-view observations are expensive to obtain in daily life scenarios.
Thus, 3D human pose estimation from monocular images/videos is highly demanded.
Some works explored to directly infer 3D human pose from monocular images/videos with end-to-end deep neural networks~\cite{pavlakos2017coarse,tekin2016structured}.
This end-to-end idea is elegant and free of accumulated errors.
However, we need paired data (images/videos and corresponding 3D poses) to supervise the training, while manually labeling 3D human poses for images/videos is impractical.
%
%
To this end, Martinez \etal.~\cite{martinez2017simple} proposed to divide 3D human pose estimation into 2D keypoint detection followed by lifting 2D joint locations to 3D positions, such that the first stage can be trained with manually labeled 2D poses and the training data for the second stage can be obtained by projecting 3D poses, obtained from motion capture devices, to 2D space.    
This divide-and-conquer strategy benefits from intermediate supervision and outperforms the end-to-end counterparts.
A family of approaches~\cite{chen20173d,hossain2018exploiting,pavllo20193d,wang2020motion,cheng2019occlusion,zeng2020srnet,ci2019optimizing,zhang2021deep} followed this strategy and focused more on lifting 2D to 3D.
Our method can also be categorized into this group.

Recently, Hossain \etal.~\cite{hossain2018exploiting} proposed to employ LSTM~\cite{hochreiter1997long} to leverage temporal information for lifting 2D joint locations to 3D positions.
Dilated temporal convolutions~\cite{pavllo20193d} and temporal attention~\cite{liu2020attention} are further explored for better temporal information aggregation.
Besides, some works focus on specific issues in the 3D human pose estimation, \eg, tackling occlusion problems~\cite{cheng2019occlusion,cheng20203d}, relieving the unreliable input issue by kinematics analysis~\cite{xu2020deep}, or addressing the lack of 3D annotations in the wild and over-fitting issues by weakly-supervised learning~\cite{iqbal2020weakly,pavllo20193d,wandt2019repnet}.
Also, Lin \etal.~\cite{lin2019trajectory} proposed to predict the 3D human pose in trajectory space by factoring the pose sequence into a trajectory base matrix and a coefficient matrix. 
The above methods represent human poses as a temporal sequence of independent joint location vectors. However, this representation cannot fully express the dependence among highly correlated human joints.


\vspace{-3.5mm}
\subsection{Graph Convolution Network (GCN)}

GCNs~\cite{scarselli2008graph,cnn_graph,gilmer2017neural} generalize conventional convolution operators to graphs, and can be roughly categorized into spectral~\cite{cnn_graph,li2018adaptive,levie2018cayleynets} and spatial~\cite{duvenaud2015convolutional,atwood2016diffusion,gilmer2017neural} perspective GCNs.
Our proposed U-CondDGCN is more related to the latter one.
Interested readers are referred to~\cite{wu2020comprehensive} for a complete survey of GCNs.

By representing human skeletons as graphs, GCNs have significantly improved a series of human-related reasoning tasks, \eg, action recognition~\cite{yan2018spatial,shi2019two,shi2019skeleton}, action synthesis~\cite{yan2019convolutional} and pose tracking~\cite{ning2020lighttrack}.
Moreover, several works~\cite{cai2019exploiting,wang2020motion,zou2020high,liu2020comprehensive,zhao2019semantic} adopt undirected graphs to represent human skeletons and apply GCN to utilize the prior knowledge of human skeleton.
However, the undirected graph representation fails to reflect the articulated characteristic of human skeletons as the hierarchical orders among joints are not explicitly presented.
In contrast, directed graph representation explicitly models hierarchical relationships among the nodes by the directions of edges.
Recently, Shi~\etal.~\cite{shi2019skeleton} also employed the directed graph representation and proposed a directed graph neural network for action recognition.
%
%
Their method learns the graph topology from the training data rather than simply defining it based on the natural structure of human skeletons.
However, the graph topology is fixed at inference time, which means different data samples still share the same topology.
%
Differently, our conditional directed graph convolution allows different poses to benefit from appropriate non-local dependence at both training and inference time, by conditioning the graph topology on input poses.
This mechanism is crucial for 3D human pose estimation since the optimal non-local dependence varies a lot for different poses, as shown in Figure~\ref{fig: conditional_connections}.


\section{Approach}
\label{sec:method}
Our full pipeline is illustrated in Figure~\ref{fig: overview}.
We first construct a temporal sequence of directed graphs from a sequence of human poses in the 2D image space $P_{\text{2D}} = \{ \bm{X}_{t,j} \in \mathbb{R}^{2} \;|\; t=1,2,3,...,T; \; j=1,2,3,...,J\}$, where $T$ and $J$ denote the number of frames in the sequence and joints on the human skeleton, respectively.
The input 2D pose sequence $P_{\text{2D}}$ can be estimated from monocular videos by any off-the-shelf 2D pose estimators, \eg, CPN~\cite{chen2018cascaded}, HR-Net~\cite{sun2019deep}, or OpenPose~\cite{cao2019openpose}. 
The nodes in the directed graph represent major joints of the human skeleton, while the edges represent the bones among the joints, as shown in Figure~\ref{fig: BVH} (b).
We set the directions of edges following the convention definition in the BVH file format.
And the ``hip'' joint, marked as the red dot in Figure~\ref{fig: BVH} (b), is set as the root node since it is the gravity center of the human body.
The features associated with nodes and edges are initialized as the joints' locations and their first-order derivatives (the difference between the child joint and parent joint), respectively.
Formally, a temporal sequence of directed graphs can be formulated as $\mathcal{G}_{\text{2D}} = \{G_t = (\mathcal{N},\; \mathcal{E}) \; | \; t=1,2,3,...,T\}$, where $\mathcal{N}$ is the set of nodes, and $\mathcal{E}$ is the set of directed edges.
We then apply our \emph{U-shaped Conditional Directed Graph Convolutional Network} (U-CondDGCN) to estimate the pose sequence in the 3D physical space $P_{\text{3D}} = \{ \bm{X}_{t,j} \in \mathbb{R}^{3} \;|\; t=1,2,3,...,T; \; j=1,2,3,...,J\}$.
%

\subsection{Network Blocks}
\label{sec:ST-CondDGConv}
After representing the 2D human pose as a sequence of directed graphs, the problem now lies in how to extract features from them to estimate the 3D pose.
To aggregate features both spatially and temporally, we use five types of blocks in our network, \ie, spatial-temporal directed graph convolution (ST-DGConv), spatial-temporal conditional directed graph convolution (ST-CondDGConv), temporal downsampling, temporal upsampling, and FC blocks, as shown in Figure~\ref{fig: overview}. 
%

\begin{figure}[!t] 
	\centering
	\includegraphics[width=0.99\linewidth]{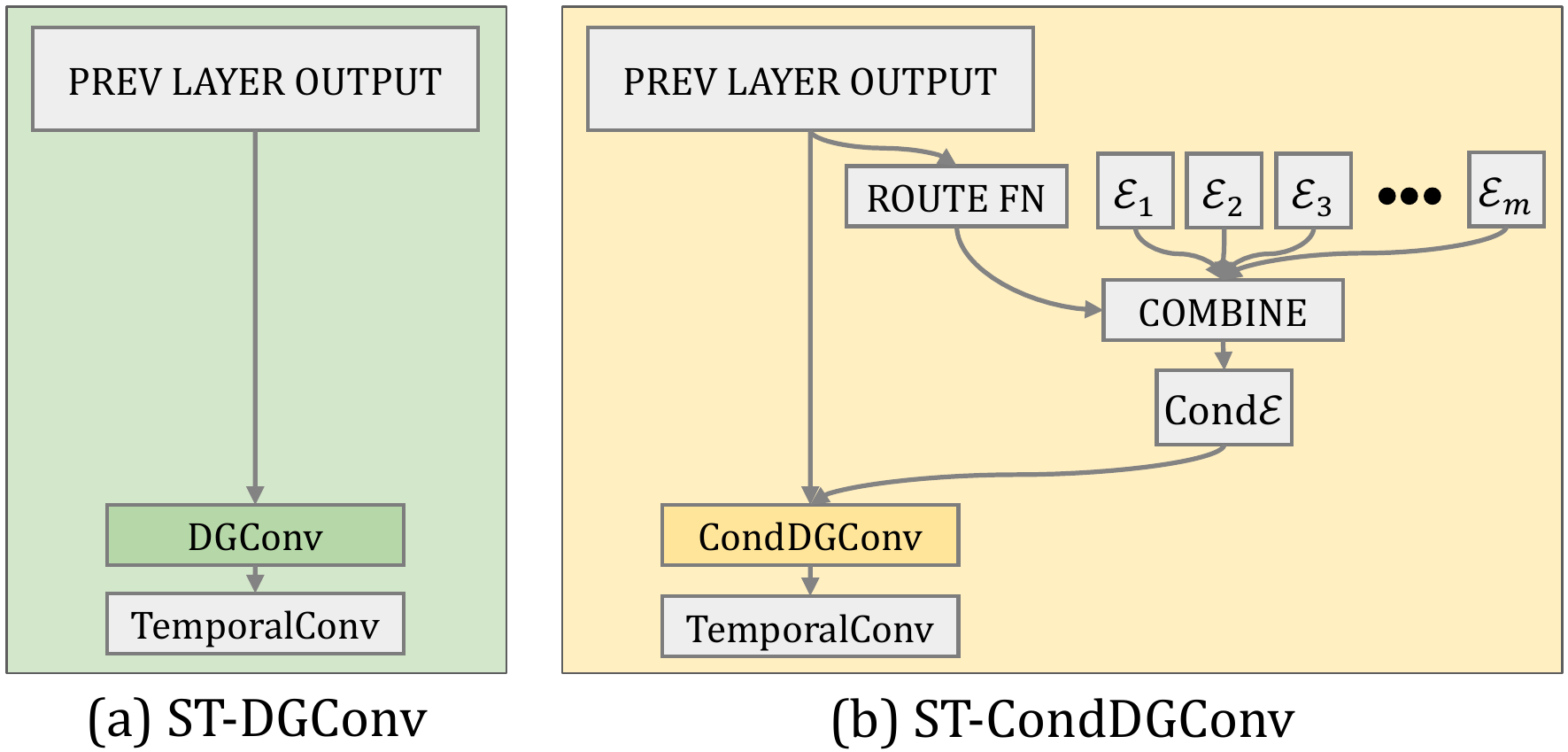}
	\vspace{-3mm}
	\caption{
		Spatial-temporal directed graph convolution (ST-DGConv) and spatial-temporal conditional directed graph convolution (ST-CondDGConv) blocks.
	}
	\vspace{-5mm}
	\label{fig: ST-CondDGConv}
\end{figure}

\paragraph{ST-DGConv}
The ST-DGConv block consists of a directed graph convolution (DGConv) followed by a temporal convolution, as shown in Figure~\ref{fig: ST-CondDGConv} (a).
DGConv exploits the spatial relationship by aggregating features from neighboring edges or nodes.
%
Details of DGConv are to be presented in Section~\ref{sec: CondDGConv}.
On the other hand, to take advantage of the temporal relationship, we employ a temporal convolution, which is a conventional 1D convolution, since the temporal sequence of directed graphs has a regular grid structure along the temporal dimension.
%

\paragraph{ST-CondDGConv}
The aforementioned ST-DGConv is based on a fixed directed graph connection $\mathcal{E}$,
which is defined according to the natural structure of human skeletons.
However, the predefined connections can only utilize the local dependence among hierarchically neighboring joints.
More importantly, sharing the same connections among all kinds of poses may not be optimal since the non-local dependence varies a lot for different poses, as shown in Figure~\ref{fig: conditional_connections}.
Inspired by the conditional convolution~\cite{NEURIPS2019_f2201f51,tian2020conditional} that enables different data samples using different convolution kernels, we propose a spatial-temporal conditional directed graph convolution (ST-CondDGConv) to condition the connections of the directed graph on input poses, such that different poses can adopt appropriate connections to exploit varied non-local dependence.
As shown in Figure~\ref{fig: ST-CondDGConv} (b), the ST-CondDGConv predicts the conditional connections $ \text{Cond}\mathcal{E}$ from previous layer's output. 
%
Specifically, there are a series of trainable connection matrix bases $\{ \mathcal{E}_1, \mathcal{E}_2, \mathcal{E}_3, ..., \mathcal{E}_m \}$, where $\mathcal{E}_{i} \in\mathbb{R}^{J \times J} $ and $m$ is the number of bases (set to $16$ in our experiments), for the network to learn.
To encourage the sparsity of the connection matrix bases, we employ the sparse initialization~\cite{martens2010deep} before the network training.  
We use a routing function to predict the blending weights for the connection matrix bases from the previous layer's output.
The network structure of the routing function is the same as that used in conditional convolution~\cite{NEURIPS2019_f2201f51}, which is a global average pooling layer followed by a fully connected layer and a Sigmoid activation.
After that, we linearly combine the bases by the predicted blending weights to produce the conditional connections $\text{Cond}\mathcal{E}$, which are then fed into the CondDGConv to aggregate the spatial information both locally and non-locally.
Note that non-local connections here are conditioned on the previous layer's output.
Thus, the routing function is able to differentiate input samples to allow different kinds of poses to leverage appropriate non-local dependence at both training and inference time.
Finally, a temporal convolution is employed to aggregate the temporal information.

Besides the above two major blocks, the other three types of blocks can be easily derived.
The temporal downsampling block is the ST-DGConv with the inside temporal convolution's stride setting to two.
It is used to downsample the temporal resolution for the larger receptive field.
The temporal upsampling block is the conventional bilinear interpolation along the temporal axis to recover higher temporal resolution.
The FC block is the standard fully-connected layer to predict the final 3D pose from the extracted features on directed graphs.

%

\subsection{Conditional Directed Graph Convolution}
\label{sec: CondDGConv}

\begin{figure*}[!t] 
	\centering
	\includegraphics[width=0.95\linewidth]{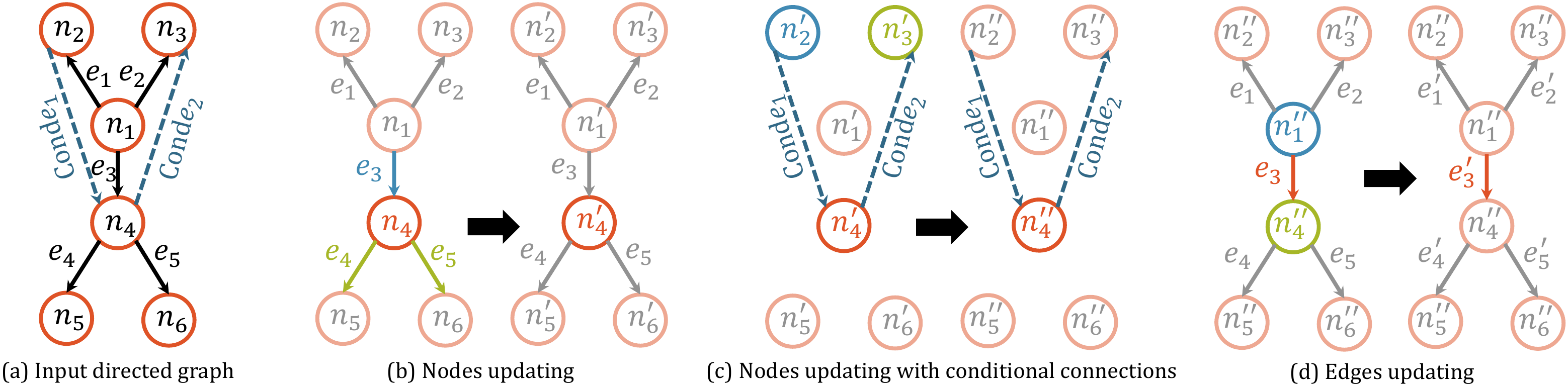}
	\vspace{-3mm}
	\caption{
		Conditional directed graph convolution (CondDGConv). Part (a) is the input directed graph with predefined connections ($e_i$) and predicted conditional connections (Cond$e_i$). Part (b), (c) and (d) are three steps of the CondDGConv.
	}
	\vspace{-3mm}
	\label{fig: CDGCN}
\end{figure*} 
In this section, we first formulate our conditional directed graph convolution (CondDGConv), which is the key operator in the ST-CondDGConv block.
The directed graph convolution (DGConv), the key operator in ST-DGConv, can be easily derived from it.
As shown in Figure~\ref{fig: CDGCN} (a), the input directed graph for CondDGConv has the predefined connections $\mathcal{E} = \{e_i \; | \; i=1, 2, ..., E\}$ and the predicted conditional connections $\text{Cond}\mathcal{E} = \{\text{Cond}e_i \; | \; i=1,2,...,C\}$, where $E$ and $C$ are the number of predefined connections and conditional connections, respectively.
The predefined connections and conditional connections are shown as arrows and blue dash-line arrows in Figure~\ref{fig: CDGCN}, respectively.
%
The CondDGConv can be formulated with three steps, as shown in Figure~\ref{fig: CDGCN} (b), (c), and (d).
%


\vspace{-2mm}
\paragraph{Nodes updating.}
Connections of the directed graph used in this step are the predefined connections, $\mathcal{E} = \{e_i \; | \; i=1, 2, ..., E\}$.
For each node $n_i$, we have the set of incoming edges that are heading into it, $\mathcal{E}^-_i$, and the set of outgoing edges that are heading out from it, $\mathcal{E}^+_i$.
Following the idea of conventional convolution, aggregating neighboring features, the nodes updating is defined as the aggregation of its incoming edge set, its outgoing edge set, and itself.
However, the number of elements in the outgoing edge set is varying. For example, as shown in Figure~\ref{fig: CDGCN} (b), the incoming edge set of node $n_4$ is $\mathcal{E}^-_4 = \{e_3\}$ while the outgoing edge set is $\mathcal{E}^+_4 = \{e_4, e_5\}$.
Therefore, we employ a pooling function to summarize features of the outgoing edge set.
Mathematically, the nodes updating step can be formulated as:
\begin{equation}
	f(n^\prime_i) = \sigma(\mathbf{w} \cdot [ f(\mathcal{E}^-_i); \;  f(n_i); \; \mathcal{P}(f(\mathcal{E}^+_i))]^\text{T} + \mathbf{b}),
	\label{eq: node_update}
\end{equation} 
where $f(\cdot)$ is the mapping from nodes/edges to their associated features; $\mathcal{P}(\cdot)$ is the pooling function (average pooling is adopted in our implementation); $\mathbf{w}$ and $\mathbf{b}$ are the trainable parameters that are similar with the kernel and bias of conventional convolution, respectively; and $\sigma$ is the activation function (ReLU is used in our implementation).

\vspace{-2mm}
\paragraph{Nodes updating with conditional connections.}
This step is designed to address the varying non-local dependence for different poses.
Thus, connections of the directed graph used in this step are the predicted conditional connections, $\text{Cond}\mathcal{E} = \{\text{Cond}e_i \; | \; i=1,2,...,C\}$.
For each node $n^{\prime}_i$, we have the parent nodes set $\mathcal{N}^{\text{p}}_i$ that have edges directed at $n^{\prime}_i$ and the child nodes set $\mathcal{N}^{\text{c}}_i$ that have edges directed from $n^{\prime}_i$.
For example, as shown in Figure~\ref{fig: CDGCN} (c), node $n^{\prime}_4$ has the parent nodes set $\mathcal{N}^{\text{p}}_4 = \{ n^{\prime}_2 \}$ and the child nodes set $\mathcal{N}^{\text{c}}_4 = \{ n^{\prime}_3 \}$.
Similarly, the nodes updating with conditional connections step can be formulated as:
\begin{equation}
	f(n^{\prime\prime}_i) = \sigma(\mathbf{w} \cdot [ \mathcal{P}(f(\mathcal{N}^p_i)); \;  f(n^{\prime}_i); \; \mathcal{P}(f(\mathcal{N}^c_i))]^\text{T} + \mathbf{b}).
\end{equation}

\vspace{-4mm}
\paragraph{Edges updating.}
Connections of the directed graph used in this step are the predefined connections, $\mathcal{E} = \{e_i \; | \; i=1, 2, ..., E\}$.
For each edge $e_i$, we have the source node $\mathcal{N}^s_i$ and the target node $\mathcal{N}^t_i$, \eg, as shown in Figure~\ref{fig: CDGCN} (d), the source node and target node of edge $e_3$ are $n^{\prime\prime}_1$ and $n^{\prime\prime}_4$, respectively.
Again, we can formulate the edges updating step as:
\begin{equation}
	f(e^{\prime}_i) = \sigma(\mathbf{w} \cdot [ f(\mathcal{N}^s_i); \;  f(e_i); \; f(\mathcal{N}^t_i)]^\text{T} + \mathbf{b}).
\end{equation}

\vspace{-2mm}
After these three steps, features associated with both nodes and edges are updated.
Note that even the edges updating step adopts the predefined connections; the non-local dependence can also be aggregated into the edges' features since they are updated by the source and target nodes' features that have aggregated the non-local information.
%
%
%
Removing step (ii), nodes updating with conditional connections, can yield the directed graph convolution (DGConv), which is purely based on the predefined connections.
Thus, it can utilize the prior knowledge of the natural structure of human skeletons but fails to leverage the varying non-local dependence for different poses. 

\subsection{Full Network}
We construct the full network as a U-shaped conditional directed graph convolutional network (UCondDGCN) using the aforementioned blocks to capture temporal relationships in both short temporal intervals and long temporal ranges.
%
As shown in the middle part of Figure~\ref{fig: overview}, the UCondDGCN has three stages: i) downsampling stage to aggregate information at long-time ranges by temporal pooling; ii) upsampling stage to recover the temporal resolution, and there are skip connections between the downsampling and upsampling stages to integrate the low-level details; iii) merging stage to combine multi-scale feature maps to predict the final 3D poses.
The numbers shown in the figure indicate the shape of features in the network, where $B$ is the batch size, $J$ is the number of nodes, $E$ is the number of edges, and $T$ is the sequence length.
As discussed in Section~\ref{sec: CondDGConv}, the ST-DGConv can better utilize the prior knowledge of the natural structure of human skeletons, whereas ST-CondDGConv allows different kinds of poses to adopt appropriate non-local dependence.
To balance the stability and flexibility, we adopt ST-DGConv in the downsampling and upsampling stages while adopting ST-CondDGConv in the merging stage.
This configuration is further justified with ablation study as shown in Table~\ref{tab: abla_cfg}.
Overall, the input of our UCondDGCN is the 2D human poses represented in directed graphs, and the output is the corresponding 3D poses, as shown in Figure~\ref{fig: overview}.

\begin{table*}[t]
	\centering
	\caption{
		Quantitative comparisons with state-of-the-art methods on Human3.6M under protocol \#1 and protocol \#2,
		where methods marked with $\dagger$ are video-based; T denotes the number of input frames; and CPN and HR-Net denote the input 2D poses are estimated by~\cite{chen2018cascaded} and~\cite{sun2019deep}, respectively. The best and second-best results are marked in bold and underlined, respectively.
	}
	\vspace{-3mm}
	\resizebox{1.0\textwidth}{!}{
		\begin{tabular}{@{}ll|ccccccccccccccc|c@{}}
			\toprule[1pt]
			\textbf{Protocol \#1}                                    &           &       Dir.       &       Disc       &       Eat        &      Greet       &      Phone       &      Photo       &       Pose       &      Purch.      &       Sit        &      SitD.       &      Smoke       &       Wait       &      WalkD.      &       Walk       &      WalkT.      &       Avg.       \\ \midrule[0.5pt]
			Martinez \etal.~\cite{martinez2017simple}                & (ICCV'17) &       51.8       &       56.2       &       58.1       &       59.0       &       69.5       &       78.4       &       55.2       &       58.1       &       74.0       &       94.6       &       62.3       &       59.1       &       65.1       &       49.5       &       52.4       &       62.9       \\
			Fang \etal.~\cite{fang2018learning}                      & (AAAI'18) &       50.1       &       54.3       &       57.0       &       57.1       &       66.6       &       73.3       &       53.4       &       55.7       &       72.8       &       88.6       &       60.3       &       57.7       &       62.7       &       47.5       &       50.6       &       60.4       \\
			Zhao \etal.~\cite{zhao2019semantic}                      & (CVPR'19) &       47.3       &       60.7       &       51.4       &       60.5       &       61.1       &      {49.9}      &       47.3       &       68.1       &       86.2       &  \textbf{55.0}   &       67.8       &       61.0       &       42.1       &       60.6       &       45.3       &       57.6       \\
			Liu \etal.~\cite{liu2020comprehensive}                   & (ECCV'20) &       46.3       &       52.2       &       47.3       &       50.7       &       55.5       &       67.1       &       49.2       &       46.0       &       60.4       &       71.1       &       51.5       &       50.1       &       54.5       &       40.3       &       43.7       &       52.4       \\ \midrule[0.5pt]
			Cai \etal.~\cite{cai2019exploiting}$^\dagger$ (CPN, T=7) & (ICCV'19) &       44.6       &       47.4       &       45.6       &       48.8       &       50.8       &       59.0       &       47.2       &       43.9       &       57.9       &       61.9       &       49.7       &       46.6       &       51.3       &       37.1       &       39.4       &       48.8       \\
			Pavllo \etal.~\cite{pavllo20193d}$^\dagger$ (CPN, T=243) & (CVPR'19) &       45.2       &       46.7       &       43.3       &       45.6       &       48.1       &       55.1       &       44.6       &       44.3       &       57.3       &       65.8       &       47.1       &       44.0       &       49.0       &       32.8       &       33.9       &       46.8       \\
			Xu \etal.~\cite{xu2020deep}$^\dagger$ (CPN, T=9)         & (CVPR'20) & \underline{37.4} &      {43.5}      &       42.7       &      {42.7}      &      {46.6}      &       59.7       &      {41.3}      &       45.1       &      {52.7}      &      {60.2}      &       45.8       &      {43.1}      &       47.7       &       33.7       &       37.1       &       45.6       \\
			Liu \etal.~\cite{liu2020attention}$^\dagger$(CPN, T=243) & (CVPR'20) &       41.8       &       44.8       &       41.1       &       44.9       &       47.4       &       54.1       &       43.4       &       42.2       &       56.2       &       63.6       &       45.3       &       43.5       &       45.3       & \underline{31.3} &       32.2       &       45.1       \\
			UGCN~\cite{wang2020motion}$^\dagger$ (CPN, T=96)         & (ECCV'20) &       41.3       &       43.9       &       44.0       &       42.2       &       48.0       &       57.1       &       42.2       &       43.2       &       57.3       &       61.3       &       47.0       &       43.5       &       47.0       &       32.6       &       31.8       &       45.6       \\
			UGCN~\cite{wang2020motion}$^\dagger$ (HR-Net, T=96)      & (ECCV'20) &      {38.2}      &  \textbf{41.0}   &      {45.9}      &      {39.7}      &  \textbf{41.4}   & \underline{51.4} &      {41.6}      & \underline{41.4} & \underline{52.0} & \underline{57.4} & \underline{41.8} &      {44.4}      & \underline{41.6} &      {33.1}      &      {30.0}      & \underline{42.6} \\ \midrule[0.5pt]
			Ours$^\dagger$ (CPN, T=96)                               &           &       38.0       &       43.3       & \underline{39.1} & \underline{39.4} &       45.8       &       53.6       & \underline{41.4} & \underline{41.4} &       55.5       &       61.9       &       44.6       & \underline{41.9} &       44.5       &       31.6       & \underline{29.4} &       43.4       \\
			Ours$^\dagger$ (HR-Net, T=96)                            &           &  \textbf{35.5}   & \underline{41.3} &  \textbf{36.6}   &  \textbf{39.1}   & \underline{42.4} &  \textbf{49.0}   &  \textbf{39.9}   &  \textbf{37.0}   &  \textbf{51.9}   &       63.3       &  \textbf{40.9}   &  \textbf{41.3}   &  \textbf{40.3}   &  \textbf{29.8}   &  \textbf{28.9}   &  \textbf{41.1}   \\ \toprule[1pt]
			\textbf{Protocol \#2}                                    &           &       Dir.       &       Disc       &       Eat        &      Greet       &      Phone       &      Photo       &       Pose       &      Purch.      &       Sit        &      SitD.       &      Smoke       &       Wait       &      WalkD.      &       Walk       &      WalkT.      &       Avg.       \\ \midrule[0.5pt]
			Martinez \etal.~\cite{martinez2017simple}                & (ICCV'17) &       39.5       &       43.2       &       46.4       &       47.0       &       51.0       &       56.0       &       41.4       &       40.6       &       56.5       &       69.4       &       49.2       &       45.0       &       49.5       &       38.0       &       43.1       &       47.7       \\
			Fang \etal.~\cite{fang2018learning}                      & (AAAI'18) &       38.2       &       41.7       &       43.7       &       44.9       &       48.5       &       55.3       &       40.2       &       38.2       &       54.5       &       64.4       &       47.2       &       44.3       &       47.3       &       36.7       &       41.7       &       45.7       \\
			Liu \etal.~\cite{liu2020comprehensive}                   & (ECCV'20) &       35.9       &       40.0       &       38.0       &       41.5       &       42.5       &       51.4       &       37.8       &       36.0       &       48.6       &       56.6       &       41.8       &       38.3       &       42.7       &       31.7       &       36.2       &       41.2       \\ \midrule[0.5pt]
			Cai \etal.~\cite{cai2019exploiting}$^\dagger$ (CPN, T=7) & (ICCV'19) &       35.7       &       37.8       &       36.9       &       40.7       &       39.6       &       45.2       &       37.4       &       34.5       &       46.9       &      {50.1}      &       40.5       &       36.1       &       41.0       &       29.6       &       33.2       &       39.0       \\
			Pavllo \etal.~\cite{pavllo20193d}$^\dagger$ (CPN, T=243) & (CVPR'19) &       34.1       &       36.1       &       34.4       &       37.2       &       36.4       &       42.2       &       34.4       &       33.6       &       45.0       &       52.5       &       37.4       &       33.8       &       37.8       &       25.6       &       27.3       &       36.5       \\
			Xu \etal.~\cite{xu2020deep}$^\dagger$ (CPN, T=9)         & (CVPR'20) &      {31.0}      &      {34.8}      &       34.7       &      {34.4}      &      {36.2}      &       43.9       &      {31.6}      &      {33.5}      & \underline{42.3} & \underline{49.0} &      {37.1}      &      {33.0}      &       39.1       &       26.9       &       31.9       &      {36.2}      \\
			Liu \etal.~\cite{liu2020attention}$^\dagger$(CPN, T=243) & (CVPR'20) &       32.3       &       35.2       &       33.3       &       35.8       &       35.9       &       41.5       &       33.2       &       32.7       &       44.6       &       50.9       &       37.0       &       32.4       &       37.0       &       25.2       &       27.2       &       35.6       \\
			UGCN~\cite{wang2020motion}$^\dagger$ (CPN, T=96)         & (ECCV'20) &       32.9       &       35.2       &       35.6       &       34.4       &       36.4       &       42.7       &       31.2       &       32.5       &       45.6       &       50.2       &       37.3       &       32.8       &       36.3       &       26.0       &       23.9       &       35.5       \\
			UGCN~\cite{wang2020motion}$^\dagger$ (HR-Net, T=96)      & (ECCV'20) & \underline{28.4} &  \textbf{32.5}   &      {34.4}      &      {32.3}      &  \textbf{32.5}   &      {40.9}      &       30.4       & \underline{29.3} &       42.6       &  \textbf{45.2}   & \underline{33.0} &       32.0       & \underline{33.2} & \underline{24.2} & \underline{22.9} & \underline{32.7} \\ \midrule[0.5pt]
			Ours$^\dagger$ (CPN, T=96)                               &           &       29.8       &       34.4       & \underline{31.9} & \underline{31.5} & \underline{35.1} & \underline{40.0} & \underline{30.3} &       30.8       &       42.6       & \underline{49.0} &       35.9       & \underline{31.8} &       35.0       &       25.7       &       23.6       &       33.8       \\
			Ours$^\dagger$ (HR-Net, T=96)                            &           &  \textbf{27.7}   & \underline{32.7} &  \textbf{29.4}   &  \textbf{31.3}   &  \textbf{32.5}   &  \textbf{37.2}   &  \textbf{29.3}   &  \textbf{28.5}   &  \textbf{39.2}   &       50.9       &  \textbf{32.9}   &  \textbf{31.4}   &  \textbf{32.1}   &  \textbf{23.6}   &  \textbf{22.8}   &  \textbf{32.1}   \\ \toprule[1pt]
		\end{tabular}
	}
	\label{table:h36m}
	\vspace{-3mm}
\end{table*}

%
\begin{table}[!t]
	\centering
	\caption{
		Results on Human3.6M with ground-truth 2D poses as input. Video-based methods are marked with $\dagger$.
	}
	\vspace{-3mm}
	\begin{tabular}{@{}ll|c@{}}
		\toprule
		Method                                        &           & MPJPE         \\ \midrule
		Martinez \etal.~\cite{martinez2017simple}     & (ICCV'17) & 45.5          \\
		Zhao \etal.~\cite{zhao2019semantic}           & (CVPR'19) & 43.8          \\
		Liu \etal.~\cite{liu2020comprehensive}        & (ECCV'20) & 37.8          \\ \midrule
		Cai \etal.~\cite{cai2019exploiting}$^\dagger$ & (ICCV'19) & 37.2          \\
		Pavllo \etal.~\cite{pavllo20193d}$^\dagger$   & (CVPR'19) & 37.2          \\
		Liu \etal.~\cite{liu2020attention}$^\dagger$  & (CVPR'20) & 34.7          \\
		UGCN~\cite{wang2020motion}$^\dagger$          & (ECCV'20) & 25.6          \\
		\midrule
		Ours$^\dagger$                                &           & \textbf{22.7} \\ \toprule
	\end{tabular}
	\label{table:h36mgt}
	\vspace{-5mm}
\end{table}

\begin{table}[!t]
	\setlength{\tabcolsep}{0.5em}
	\centering
	\caption {Results on MPI-INF-3DHP with three metrics, where PCK and AUC are the higher, the better, while MPJPE is the lower, the better.}
	\label{tab: mpii3}
	\vspace{-3mm}
	\begin{tabular}{@{}l|ccc@{}}
		\toprule
		Method                                    & PCK[$\uparrow$] & AUC[$\uparrow$] & MPJPE[$\downarrow$] \\
		\midrule
		Mehta \etal.~\cite{mehta2017monocular}    & 75.7            & 39.3            & -                   \\
		VNect (ResNet50)~\cite{mehta2017vnect}    & 77.8            & 41.0            & -                   \\
		VNect (ResNet101)~\cite{mehta2017vnect}   & 79.4            & 41.6            & -                   \\
		TrajectoryPose3D~\cite{lin2019trajectory} & 83.6            & 51.4            & 79.8                \\
		UGCN~\cite{wang2020motion}                & {86.9}          & {62.1}          & {68.1}              \\
		\midrule
		Ours                                      & \textbf{97.9}   & \textbf{69.5}   & \textbf{42.5}       \\
		\bottomrule
	\end{tabular}
	\vspace{-5mm}
\end{table}

\section{Experiments}
\label{sec:experiment}
%
We formulate the loss function similar to UGCN~\cite{wang2020motion}, 
\begin{equation}
	\mathcal{L} = \mathcal{L}_p + \lambda \mathcal{L}_m,
\end{equation}
where $\mathcal{L}_p$ is 3D joint position loss that is known as the mean per-joint position error (MPJPE), $\mathcal{L}_m$ is motion loss~\cite{wang2020motion}, and $\lambda$ is a weight to balance them (set to $0.1$ in our implementation). 
The motion loss is originally proposed in UGCN~\cite{wang2020motion} to supervise the temporal structure of the predicted pose sequence. It first encodes the positions of the same joint at two different temporal instants into pairwise motion encodings for the predicted pose sequence and the ground truth, respectively; and then computes the L1 loss between them. 
The number of layers, kernel size, and input sequence length are also followed UGCN~\cite{wang2020motion}.

We implemented UCondDGCN on the PyTorch platform~\cite{paszke2019pytorch} and conducted experiments on a single NVIDIA TITAN V GPU.
We optimized the model by the AdaMod optimizer~\cite{ding2019adaptive} for 110 epochs with a batch size of 256, in which the learning rate was initially set to $5\times10^{-3}$ and decayed by $0.1$ after the $80^{\text{th}}$, $90^\text{th}$, and $100^{\text{th}}$ epochs.
To avoid over-fitting, we set the weight decay factor to $10^{-5}$ and the dropout rate to $0.3$.
We followed UGCN~\cite{wang2020motion} to adopt the sliding window algorithm with a step length of five frames to estimate variable sequence length at inference time.

\vspace{-2mm}
\subsection{Dataset and Evaluation Metrics}
\paragraph{Human3.6M}
Human3.6M~\cite{ionescu2013human3} is the most widely used evaluation benchmark, containing $3.6$ million video frames captured from four synchronized cameras with different locations and poses at $50$ Hz.
There are $11$ subjects performing $15$ kinds of actions, \eg, ``walking'', ``sitting'', and ``eating''.
Following previous works~\cite{pavlakos2017coarse,tekin2016direct,martinez2017simple,pavllo20193d,sun2017compositional,wang2020motion,cheng2019occlusion}, we adopted the 17-joint pose, trained a single model on five subjects (S1, S5, S6, S7, S8) for all kinds of actions, and tested it on the remaining two subjects (S9 and S11).

\vspace{-2mm}
\paragraph{MPI-INF-3DHP}
MPI-INF-3DHP~\cite{mehta2017monocular} is a relatively new dataset captured in both indoor and outdoor environments.
Similar to Human3.6M, it contains various subjects, actions, and camera settings, and
we followed~\cite{mehta2017vnect,wang2020motion,lin2019trajectory} to split the training and testing set.

\vspace{-2mm}
\paragraph{Evaluation Metrics}
For Human3.6M, we adopt the most widely used two metrics: \emph{Protocol 1} is the mean per-joint position error (MPJPE) that is the mean Euclidean distance between the estimated joint positions and ground truth in millimeters; and \emph{Protocol 2} is the error after alignment with the ground truth in translation, rotation, and scale (P-MPJPE).
For MPI-INF-3DHP, we also report the percentage of correct keypoints (PCK)~\cite{mehta2017monocular} score with the threshold of $150mm$ and the area under the curve (AUC)~\cite{mehta2017monocular} of the PCK scores with different error thresholds, following~\cite{lin2019trajectory,wang2020motion,mehta2017monocular,mehta2017vnect}.

\begin{figure*}[!t] 
	\centering
	\includegraphics[width=0.96\linewidth]{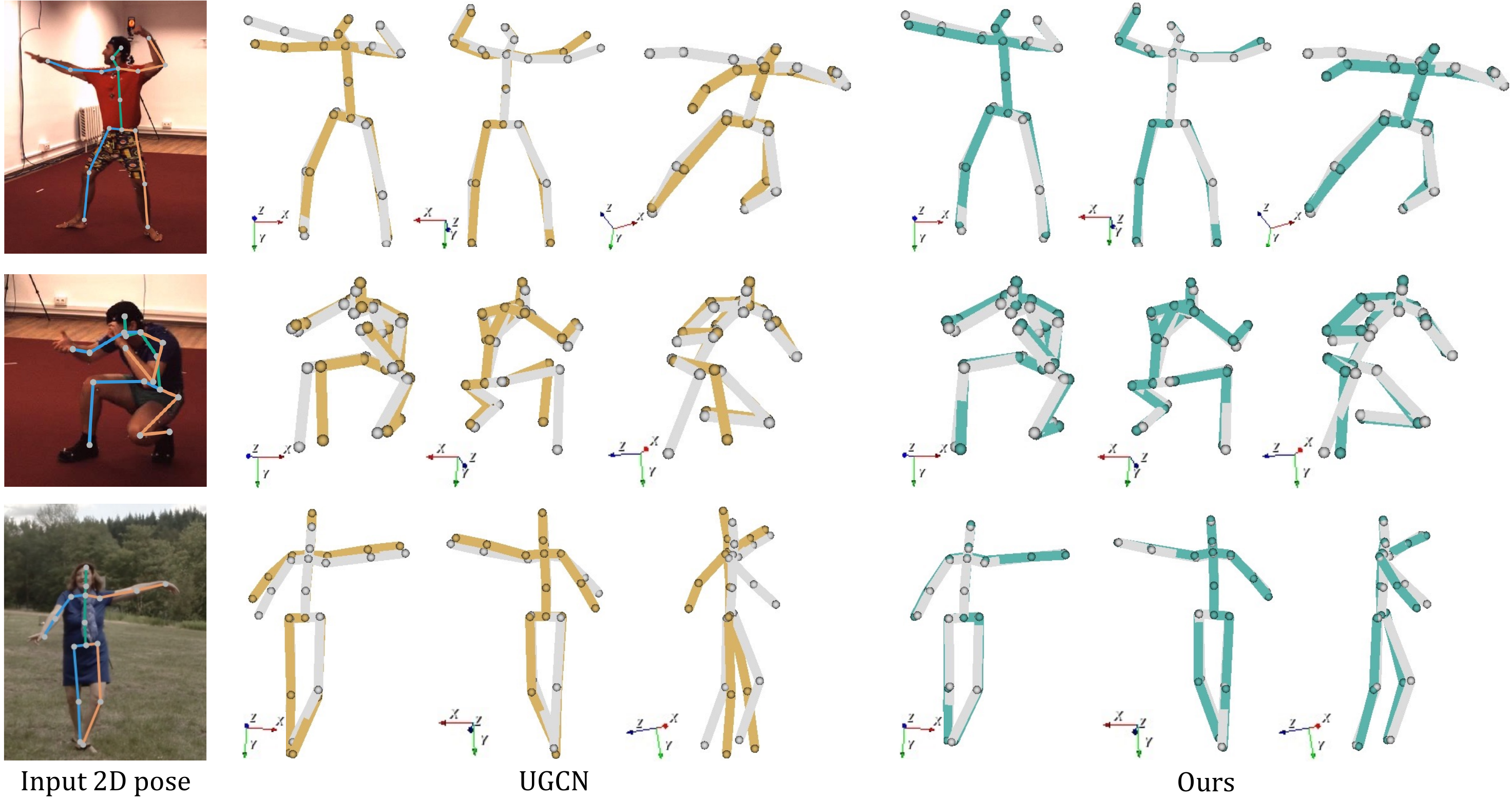}
	\vspace{-3mm}
	\caption{
		Qualitative comparison between and the baseline, \textcolor{UGCN}{UGCN}~\cite{wang2020motion} and \textcolor{OURS}{our method} on Human3.6M and MPI-INF-3DHP.
		To better evaluate 3D poses' quality, we show them under three different viewpoints as indicated by 3D orientation markers.
		And ground-truth 3D poses are shown in \textcolor{GT}{gray} as a reference.
	}
	\vspace{-3mm}
	\label{fig: visual}
\end{figure*} 

\vspace{-1mm}
\subsection{Quantitative Evaluation}
\paragraph{Results on Human3.6M}
To evaluate the effectiveness of our U-CondDGCN, we first quantitatively compared our method with state-of-the-art methods on the Human3.6M benchmark in Table~\ref{table:h36m}, including image-based methods~\cite{martinez2017simple,fang2018learning,zhao2019semantic,liu2020comprehensive} and video-based methods~\cite{cai2019exploiting,pavllo20193d,xu2020deep,liu2020attention,wang2020motion} (marked with $\dagger$).
The input 2D poses are estimated from monocular images/videos by either CPN~\cite{chen2018cascaded} or HR-Net~\cite{sun2019deep} (a more powerful 2D pose estimator).
We can see that video-based methods generally perform better than image-based methods, which justifies that the temporal information is beneficial to the 3D human pose estimation.
More importantly, from both Protocol \#1 and Protocol \#2 results, we can see our method consistently outperforms all the others by a large margin no matter with CPN or HR-Net input 2D poses (\emph{1.7mm} and \emph{1.5mm} improvements in terms of Protocol \#1, respectively).
%
%
This demonstrates the effectiveness of our proposed conditional directed graph convolution.

To further explore the upper bound of our U-CondDGCN for lifting 2D poses to 3D poses, we compared our method with several state-of-the-art methods with ground-truth 2D poses as input since doing so can eliminate the influence of the 2D pose estimators applied.
As shown in Table~\ref{table:h36mgt}, we can see our method significantly outperforms all the others ($\geq2.9mm$) in terms of MPJPE.
It demonstrates if a more powerful 2D pose estimator is available, our U-CondDGCN is able to produce more accurate 3D poses.

\vspace{-2mm}
\paragraph{Results on MPI-INF-3DHP}
To evaluate the generalization ability, we compared our method with state-of-the-art methods on the MPI-INF-3DHP benchmark, as shown in Table~\ref{tab: mpii3}.
We followed the experimental setting in~\cite{lin2019trajectory,wang2020motion} to adopt ground-truth 2D poses as input.
Our method achieves significant improvements no matter in terms of PCK ($11.0\%$ improvement), AUC ($7.4\%$ improvement), or MPJPE ($25.6mm$ improvement), which demonstrates our method generalizes well on various datasets.

\begin{table}[t]
	\setlength{\tabcolsep}{1em}
	\centering
	\caption {Ablation study. We compared results of the baseline (UGCN~\cite{wang2020motion}), the variant of our method (UDGCN), our U-CondDGCN, and different configurations for our U-CondDGCN on Human3.6M. The $\Delta$ denotes the improvements compared with the baseline.}
	\label{tab: abla_cfg}
	\vspace{-3mm}
	\begin{tabular}{@{}lccc@{}}
		\toprule
		Method     & Cond$\mathcal{E}$  & MPJPE           & $\Delta$       \\ \midrule
		UGCN       & -                  & 25.6            & -              \\
		UDGCN      & -                  & 23.9            & 1.7            \\
		U-CondDGCN & Merging stage      & \textbf{{22.7}} & \textbf{{2.9}} \\ 
		\midrule
		U-CondDGCN & Upsampling stage   & {24.1}          & {1.5}          \\
		U-CondDGCN & Downsampling stage & {25.3}          & {0.3}          \\
		U-CondDGCN & All stages         & {23.6}          & {2.0}          \\ 
		\bottomrule
	\end{tabular}
	\vspace{-5mm}
\end{table}

\vspace{-3mm}
\subsection{Qualitative Evaluation}
To qualitatively evaluate our U-CondDGCN, we compared it against the baseline, UGCN~\cite{wang2020motion}.
As shown in Figure~\ref{fig: visual}, we show several example results of UGCN and ours under various viewpoints in orange and teal, respectively, while the ground truth 3D poses are shown in neutral gray as a reference.
The first two examples are from Human3.6M, and the last example is from MPI-INF-3DHP.
The input 2D poses for Human3.6M are estimated from monocular videos by the 2D pose estimator, HR-Net~\cite{sun2019deep}, while that for MPI-INF-3DHP is the ground-truth 2D poses.
We can see our results are more consistent with the ground-truth 3D poses, especially for the end-effectors, \eg, the ``wrist'' and ``ankle'' joints.
It qualitatively demonstrates the effectiveness of our U-CondDCCN for estimating 3D poses.
Readers are highly recommended to watch the supplementary video to better explore the performance.

\vspace{-1mm}
\subsection{Analysis}

\paragraph{Effectiveness of CondDGConv}

To explore the effectiveness of directed graph representation and our proposed CondDGConv, we conducted ablation experiments on the Human3.6M dataset by considering the following methods:
\vspace{-1mm}
\begin{itemize}
	\item UGCN~\cite{wang2020motion} is the baseline, which adopts the undirected graph representation and has the same U-shaped structure;
	\item UDGCN is a variant of our method that replaces all the CondDGConv with the DGConv; and
	\item U-CondDGCN is the full version of our method.
	\vspace{-1mm}
\end{itemize}
We adopted the ground-truth 2D poses as input to eliminate the influence of the 2D pose estimator.
The results are shown in the top part of Table~\ref{tab: abla_cfg}.
Comparing the results of UGCN and UDGCN, we can see that UDGCN outperforms the UGCN by $1.7mm$ in terms of MPJPE.
It shows the directed graph can better model the hierarchy of the articulated human skeleton, and the hierarchy is beneficial to the 3D pose reasoning.
Moreover, comparing the results of UDGCN and U-CondDGCN, we can find CondDGConv can further improve the performance by $1.2mm$.
It demonstrates adaptive graph topologies for different pose samples can better leverage the varying non-local dependence.

\begin{figure}[!t] 
	\centering
	\includegraphics[width=0.97\linewidth]{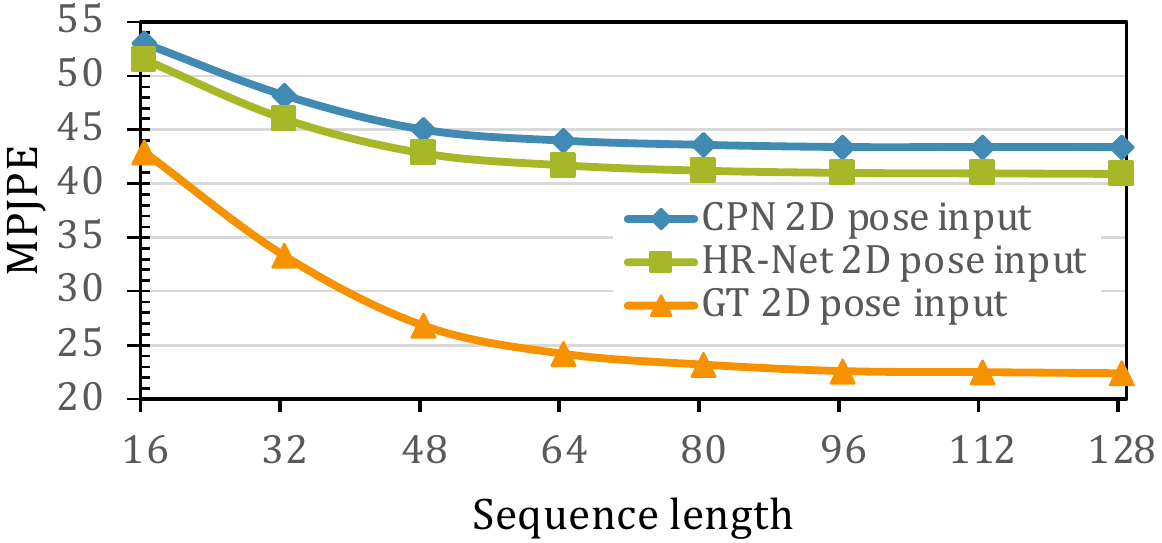}
	\vspace{-3mm}
	\caption{
		Influence of the input sequence length. Curves show the quality of estimated 3D poses against the input sequence length with the input 2D poses from CPN~\cite{chen2018cascaded}, HR-Net~\cite{sun2019deep}, and ground-truth, respectively.
	}
	\vspace{-4mm}
	\label{fig: seqLen}
\end{figure} 

\vspace{-1mm}
\paragraph{Configuration of U-CondDGCN}
To explore the best stage to utilize the conditional connections (Cond$\mathcal{E}$), we conducted experiments by adopting Cond$\mathcal{E}$ at various stages, \ie, downsampling stage, upsampling stage, merging stage, and all stages, and show the results in the bottom part of Table~\ref{tab: abla_cfg}.
We can see that adopting Cond$\mathcal{E}$ at the downsampling or upsampling stage would decrease the performance compared with the UDGCN, whereas adopting the Cond$\mathcal{E}$ at the merging stage can improve the performance.
This is because introducing Cond$\mathcal{E}$ at too early stages would prevent the network from utilizing the prior knowledge of the natural structure of human skeletons while introducing Cond$\mathcal{E}$ at the late stage, \ie, the merging stage, can both keep the prior knowledge and allow different kinds of poses to adopt optimal non-local dependence.
Compared with adopting Cond$\mathcal{E}$ at all the stages (yields $2.0mm$ improvement), adopting it at the merging stage can better balance the stability and flexibility (yields $2.9mm$ improvement).



\vspace{-1mm}
\paragraph{Influence of the input sequence length}
To explore it, we measured the MPJPE of estimated 3D poses under various lengths of the input sequence with the input 2D poses from CPN~\cite{chen2018cascaded}, HR-Net~\cite{sun2019deep}, and ground-truth, and plotted curves of the MPJPE against the input sequence length in Figure~\ref{fig: seqLen}.
We can see that no matter with what kinds of input 2D poses, the performance always goes better when the sequence length increases.
It shows longer sequence can provide more temporal information for the 3D pose reasoning.
%

\begin{figure}[!t] 
	\centering
	\includegraphics[width=0.95\linewidth]{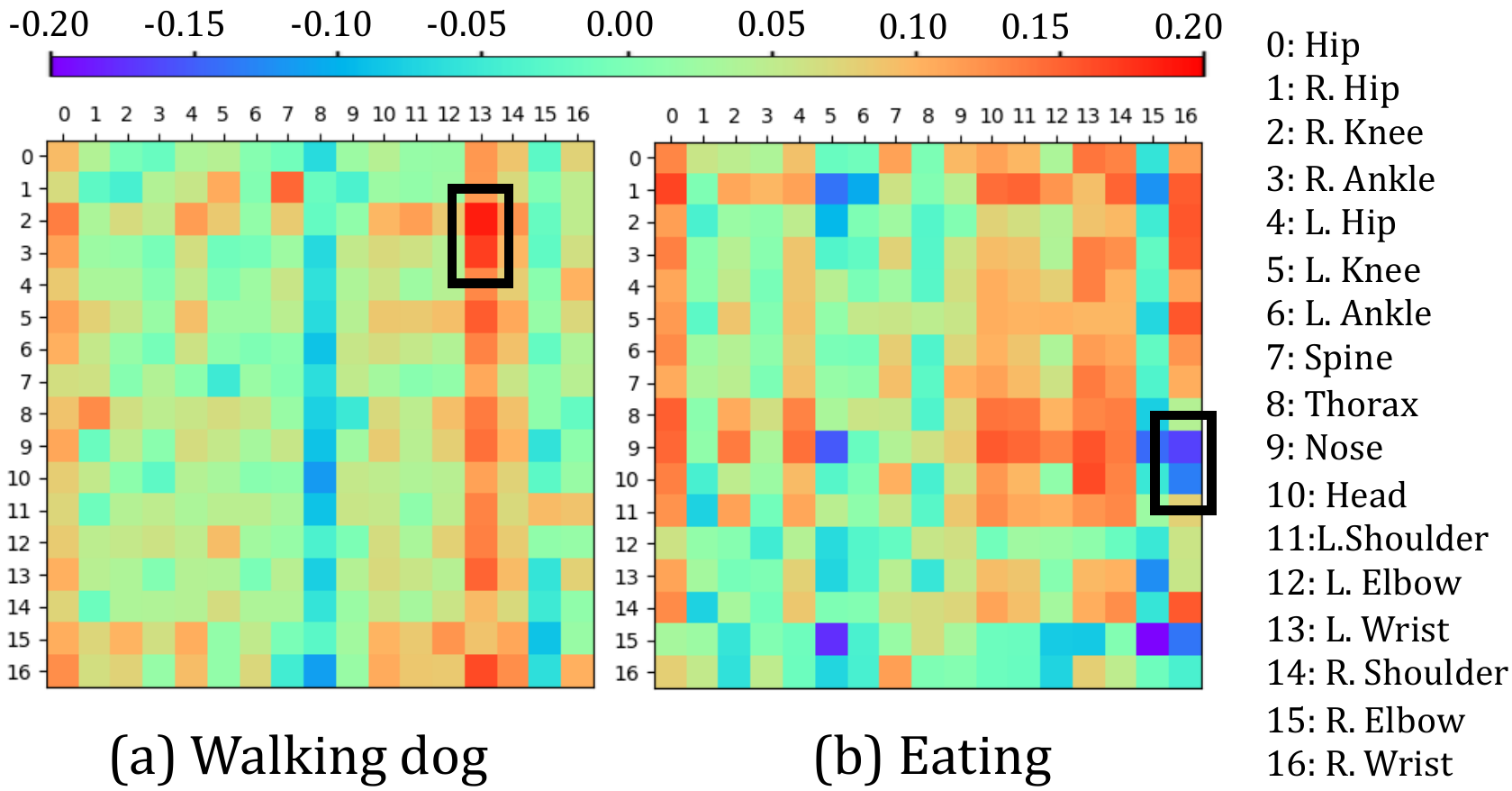}
	\vspace{-3mm}
	\caption{
		Visualization of the predicted conditional connection matrices from two kinds of actions. Two significant dependency examples are marked in black rectangles.
	}
	\vspace{-4mm}
	\label{fig: vis_map}
\end{figure}

\vspace{-1mm}
\paragraph{Visualization of predicted conditional connections.}
To verify the predicted conditional connections for different input poses, we visualized the connection matrices (from the last layer of U-CondDGCN) predicted for two pose sequences in the ``walking dog'' and ``eating'' action classes, respectively.
The connection matrix Cond$\mathcal{E}$ has a shape of $J \times J$, where $J$ is the number of nodes.
The absolute value of entry Cond$\mathcal{E}_{i,j}$ means the extent of dependence between node $i$ and $j$, while the positive/negative sign stands for node $i$ belongs to the child/parent set of node $j$, where $i$ and $j$ are the row and column indices, respectively.
%
As shown in Figure~\ref{fig: vis_map}, we can see these two connection matrices are very different.
For example, the dependency between the left wrist and the right knee/ankle nodes is significant for the ``walking dog'' sequence, whereas the dependency between the right wrist and the nose/head nodes is significant for the ``eating'' sequence, as marked with black rectangles. 
It demonstrates our U-CondDGCN can differentiate input poses to predict appropriate non-local connections at inference time.

\vspace{-1mm}
\paragraph{Inference speed.}
The number of parameters of our method is 3.42M, while that of UGCN~\cite{wang2020motion} is 1.69M. The difference mainly comes from that conventional GCN only considers nodes’ features while our CondDGConv considers both nodes’ and edges’ features. However, such a number of parameters is still far less than that of temporal-convolution-based methods such as VideoPose3D~\cite{pavllo20193d}, which has 16.95M parameters.
Under the experimental setting stated in Sec.~\ref{sec:experiment}, our method takes around 3.6ms to estimate a 3D pose from the 2D poses.


	\section{Conclusion}
	We present the conditional directed graph convolution for 3D human pose estimation from monocular videos.
	The employed directed graph representation can better model the articulated hierarchy of human skeletons than the undirected graph.
	Moreover, we present a novel conditional connection mechanism for the directed graph convolution, such that different kinds of poses can adopt appropriate non-local dependence to facilitate the 3D pose reasoning.
	Extensive quantitative and qualitative evaluations demonstrate that our method achieves top performance on two large-scale challenging benchmarks, Human3.6M and MPI-INF-3DHP.
	Also, we conducted a wide spectrum of analyses to verify our method.
	%
	We believe the insight behind the conditional directed graph convolution can also benefit other tasks where the articulated structure is involved, like the 3D hand gesture estimation and recognition.
	
\newpage
	
\vfill\eject
\bibliographystyle{ACM-Reference-Format}
\bibliography{reference}

\end{document}